%% file: iclr2020_conference.tex
\documentclass{article} 
\usepackage{iclr2020_conference,times}

\input{math_commands.tex}

\usepackage{hyperref}
\usepackage{url}
\usepackage{amsthm}
\usepackage{graphicx}
\usepackage{subfigure}

\theoremstyle{definition}

\title{Probabilistic Contrastive Loss \\for Self-Supervised Learning}

\author{Shen Li$^{\P}$\quad Jianqing Xu$^{\ddagger}$\quad Bryan Hooi$^{\P}$\\
$^{\P}$National University of Singapore\quad $^{\ddagger}$Youtu Lab, Tencent \\
\tt\small{shen.li@u.nus.edu \quad joejqxu@tencent.com \quad bhooi@comp.nus.edu.sg} \\
}

\iclrfinalcopy 
\begin{document}

\maketitle

\begin{abstract}
This paper proposes a probabilistic contrastive loss function for self-supervised learning. The well-known contrastive loss is deterministic and involves a temperature hyperparameter that scales the inner product between two normed feature embeddings. By reinterpreting the temperature hyperparameter as a quantity related to the radius of the hypersphere, we derive a new loss function that involves a confidence measure which quantifies uncertainty in a mathematically grounding manner. Some intriguing properties of the proposed loss function are empirically demonstrated, which agree with human-like predictions. We believe the present work brings up a new prospective to the area of contrastive learning.
\end{abstract}

\section{Introduction}
Self-supervised learning has witnessed a surge of research interest in contrastive loss~\citep{he2020momentum, chen2020simple} as the \emph{de facto} learning paradigm. \cite{wang2020understanding} analysed its success in terms of alignment (closeness) of features from positive pairs and uniformity of the introduced distribution of the normalized features on the hypersphere. However, the treatment is deterministic, which does not allow for uncertainty quantification that is otherwise critically important. \footnote{Work in progress. Shen Li is jointly sponsored by IDS, NUS and Google PhD fellowship.}

Uncertainty-aware representation learning can be achieved by virtue of probabilistic approaches, which serve as a natural tool for modelling data uncertainty. We draw on the recent advancement on probabilistic face recognition \citep{li2021spherical} and show that a probabilistic contrastive loss function can be derived for uncertainty-aware self-supervised learning.

\section{Contrastive Loss Revisited}
Learning from unlabeled data is an ultimate goal of representation learning. Recent literature has seen the wide usage of contrastive learning as a promising avenue to this goal. Generally, the learning process proceeds by drawing positive and negative pairs from data to contrast. Practically, the positive pairs are obtained by taking two different augmented views of the same sample and the negative pairs can be constructed by selecting views from different samples.

We follow~\citep{wang2020understanding} to use the same set of notations for development. Let $f: \mathcal{X} \mapsto \mathbb{S}^{d-1}$ denote an encoder that maps data to a $d$-dimensional hypersphere, and let $p_{\text{data}}(\cdot)$ be the data distribution over $\mathcal{X}$ and $p_{\text{pos}}(\cdot, \cdot)$ the distribution of positive pairs over $\mathcal{X} \times \mathcal{X}$. Then the encoder can be trained by minimizing the following contrastive loss function:
\begin{equation}
\label{eq:con_loss}
\mathcal{L}_{\text {contrastive }}(f ; \tau, M) := 
\underset{(x_i, x_j) \sim p_{\text {pos }}, \{x_{i}^{-}\}_{i=1}^{M} \stackrel{\text { iid }}{\sim} p_{\text {data}}}{\mathbb{E}}\left[-\log \frac{e^{f(x_i)^{\top} f(x_j) / \tau}}{e^{f(x_i)^{\top} f(x_j) / \tau}+\sum_{i} e^{f\left(x_{i}^{-}\right)^{\top} f(x_j) / \tau}}\right]
\end{equation}
where $\tau > 0$ is a scalar temperature hyperparameter, and $M$ is a fixed number of negative samples.

In our work, we reinterpret $\tau$ as a quantity related to the radius of the hypersphere by recognizing
\begin{equation}
    r = \sqrt{\frac{1}{\tau}}
\end{equation}
This leads to the following formulation:
\begin{equation}
\label{eq:con_s_loss}
    \mathcal{L}(f ; r, M) = \underset{(x_i, x_j) \sim p_{\text {pos }}, \{x_{i}^{-}\}_{i=1}^{M} \stackrel{\text { iid }}{\sim} p_{\text {data}}}{\mathbb{E}}\left[-\log \frac{e^{s(x_i, x_j)}}{e^{s(x_i, x_j)}+\sum_{i} e^{s(x_{i}^{-}, x_j)}}\right]
\end{equation}
where $s(x_i, x_j)$ is the inner product that measures the similarity between two features $z_i = rf(x_i)$ and $z_j = rf(x_j)$ residing on the $r$-radius hypersphere $r\mathbb{S}^{d-1}$.

\begin{figure}[t]
\begin{center}
  \includegraphics[width=0.6\linewidth]{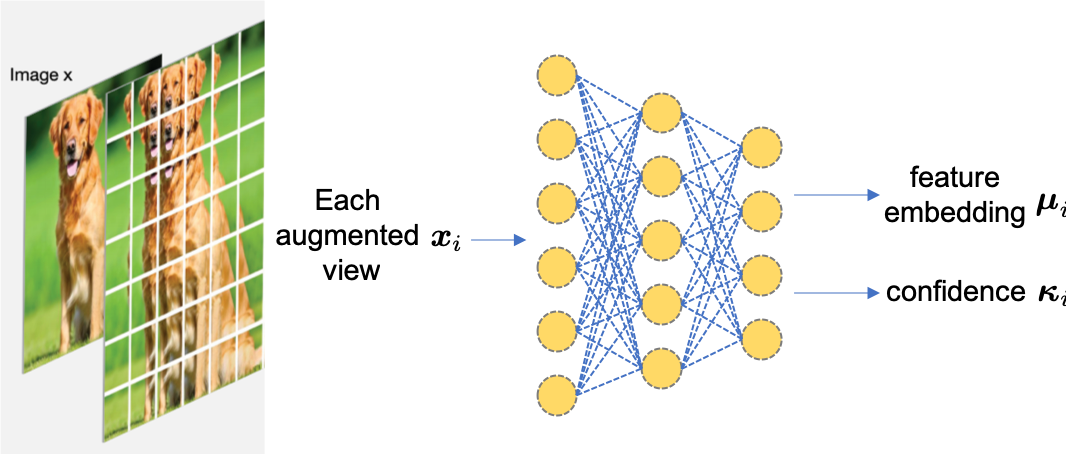}
\end{center}
  \caption{\small The schema of the proposed training procedure. Each augmented view can be assigned with confidence besides the feature embedding. The rationale behind it is that there is no reason to believe a patch of background is part of a dog.}
\label{fig:schema}
\end{figure}

\section{The Proposed Formulation}
Rewriting Eq. (\ref{eq:con_loss}) into Eq. (\ref{eq:con_s_loss}) opens up more possibilities for similarity measures beyond inner product. Instead of assuming the deterministic embeddings $z_i = rf(x_i)$, we take a probabilistic approach by treating $z_i$ as a random variable, i.e., $z_i \sim r\text{-vMF}(\mu_i(x_i), \kappa_i(x_i))$. Here, $r\text{-vMF}(\mu_i(x_i), \kappa_i(x_i))$ denotes the $r$-radius \emph{von Mises Fisher} distribution \citep{li2021spherical} with the mean direction $r\mu_i(x_i)$ and the concentration value $\kappa_i(x_i)$. Then, the similarity measure $s(\cdot, \cdot)$ can be instantiated using mutual likelihood score defined in the $r$-radius hypersphere \citep{li2021spherical}. Mutual likelihood score involves confidence measure $\kappa_i$ that naturally admits uncertainty quantification of the augmented view, $x_i$. This is expected to yield better representations for downstream tasks since different augmented views should be assigned with different confidence; for example, there is no reason to believe a patch of background is part of a dog (cf. Figure~\ref{fig:schema}). 

Mathematically, the mutual likelihood score $s(\cdot, \cdot)$ is defined as
\begin{align}
\label{eq:mls}
s({x}_i, {x}_j)
&:= \log \iint_{{r\mathbb{S}^{d-1}} \times {r\mathbb{S}^{d-1}}} p({z}_i|{x}_i)p({z}_j|{x}_j)\delta({z}_i-{z}_j)d{{z}_i}d{{z}_j} \\
&= \log\mathcal{C}_d(\kappa_i) + \log\mathcal{C}_d(\kappa_j) - \log\mathcal{C}_d{(\Tilde{\kappa})} - d\log r \\
&= \left(\frac{d}{2}-1\right)\log\left(\frac{\kappa_i \kappa_j}{\Tilde{\kappa}}\right) + \log\left(\frac{\mathcal{I}_{d/2-1}(\Tilde{\kappa})}{\mathcal{I}_{d/2-1}(\kappa_i) \cdot \mathcal{I}_{d/2-1}(\kappa_j)}\right) - d\log(\sqrt{2\pi}r)
\end{align}
where $\Tilde{\kappa}=||\kappa_i {\mu}_i + \kappa_j {\mu}_j||_2$.

Hence, the encoder $f$ can be trained by minimizing Eq.~(\ref{eq:con_s_loss}) with $s(\cdot, \cdot)$ defined in Eq.~(6). Figure~\ref{fig:schema} shows the schema of our proposed approach.

Next, we show the functional landscape of $s$ to understand how the proposed formulation operates. Note that $s$ is agnostic about the absolute position of either $\mu_i$ or $\mu_j$; instead, it depends on the relative position which can simply be quantified by the cosine distance $\cos\theta := \mu_i^T\mu_j / (\left\Vert \mu_i \right\Vert \left\Vert \mu_j \right\Vert)$. Formally, $s$ can be rewritten as a function of $\kappa_i$, $\kappa_j$ and $\cos\theta$:
\begin{equation}
s = \left(\frac{d}{2}-1\right)\log\left(\frac{\kappa_i \cdot \kappa_j}{\sqrt{\kappa_i^2 + \kappa_j^2 + 2\kappa_i\kappa_j\cos\theta}}\right) + \log \left(\frac{\mathcal{I}_{\frac{d}{2}-1}\left(\sqrt{\kappa_i^2 + \kappa_j^2 + 2\kappa_i\kappa_j\cos\theta}\right)}{\mathcal{I}_{\frac{d}{2}-1}(\kappa_i) \cdot \mathcal{I}_{\frac{d}{2}-1}(\kappa_j)}\right) - d\log(\sqrt{2\pi}r)
\end{equation}

\begin{figure}[t]
\begin{center}
  \includegraphics[width=\linewidth]{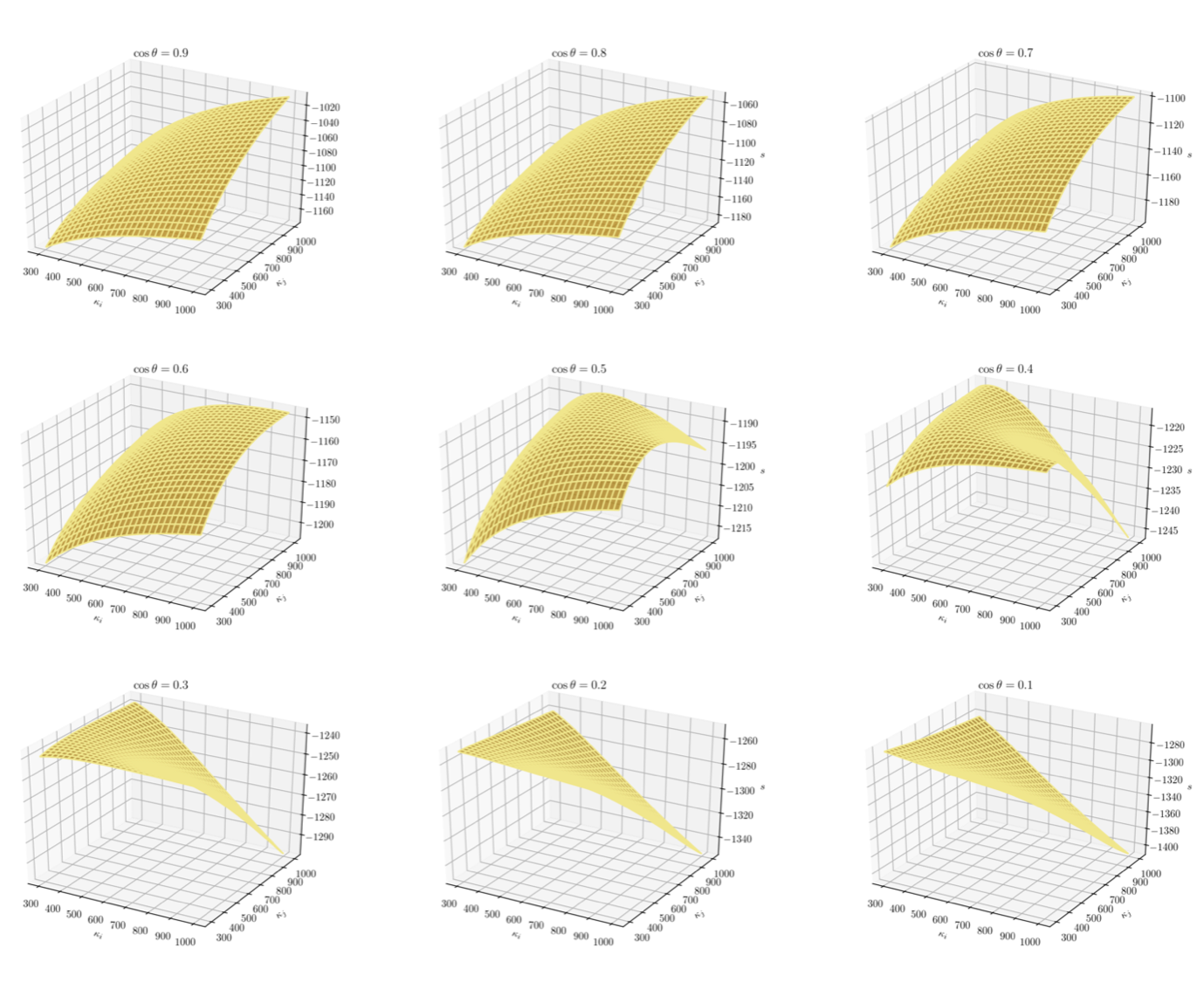}
\end{center}
  \caption{\small The function $s$ is a function of $\kappa_i$, $\kappa_j$ and $\cos\theta$.}
\label{fig:s}
\end{figure}

Figure~\ref{fig:s} demonstrates how $s$ varies according to $\kappa_i$, $\kappa_j$ and $\cos\theta$.

We observe that when $\cos\theta$ is large ($\ge 0.6$), large magnitudes of $\kappa_i$ and $\kappa_j$ yield a higher mutual likelihood score. This is a desirable effect since similar deteriministic predictions with high confidence should give rise to a higher similarity score. When $\cos\theta$ is small ($\le 0.5$), large magnitudes of $\kappa_i$ and $\kappa_j$ do not yield a high mutual likelihood score; rather, the score is even smaller than those with a high $\kappa_i$ and a low $\kappa_j$ (or the other way around). This is also a manifestation of human-like predictions, as when predictions disagree, the similarity score should be small even when confidence scores are high for both of the predictions.

\bibliography{iclr2020_conference}
\bibliographystyle{iclr2020_conference}

\end{document}

%% file: math_commands.tex
\usepackage{amsmath,amsfonts,bm}

\def\eqref#1{equation~\ref{#1}}

\def\1{\bm{1}}

\DeclareMathAlphabet{\mathsfit}{\encodingdefault}{\sfdefault}{m}{sl}
\SetMathAlphabet{\mathsfit}{bold}{\encodingdefault}{\sfdefault}{bx}{n}

